\documentclass[11pt, a4paper, logo, twocolumn, copyright]{googledeepmind}

\usepackage[authoryear, sort&compress, round]{natbib}
\title{Perception Test 2023: A Summary of the First Challenge And Outcome}

\correspondingauthor{viorica@google.com}


\author[1]{Joseph Heyward}
\author[1]{Jo\~ao Carreira}
\author[1,2]{Dima Damen}
\author[1,3]{Andrew Zisserman}
\author[1]{Viorica P\u atr\u aucean}

\affil[1]{Google DeepMind}
\affil[2]{University of Bristol}
\affil[3]{University of Oxford}

\begin{abstract}
The First Perception Test challenge was held as a half-day workshop alongside the IEEE/CVF International Conference on Computer Vision (ICCV) 2023, with the goal of benchmarking state-of-the-art video models on the recently proposed Perception Test benchmark.
The challenge had six tracks covering low-level and high-level tasks, with both a language and non-language interface, across video, audio, and text modalities, and covering: object tracking, point tracking, temporal action localization, temporal sound localization, multiple-choice video question-answering, and grounded video question-answering. We summarise in this report the task descriptions, metrics, baselines, and results.
\end{abstract}

\begin{document}

\maketitle

\section{Introduction}
Making progress towards building perception systems that achieve human-level scene understanding requires robust and comprehensive evaluations to guide research. Following-up on our previous ECCV 2022 workshop on designing and evaluating computer perception systems\footnote{\url{https://computerperception.github.io/}}, we organised this workshop-challenge to focus on evaluating SOTA perception models. 

% ~\cite{cope2022}

\noindent \textbf{Dataset:} Most of the existing benchmarks have been designed to evaluate specialised models, e.g.\ image or video classifiers, objects detectors and so on. Using a collection of such benchmarks to evaluate general perception models results in expensive evaluations that are redundant in some areas (e.g.\ semantics), while missing other areas  (e.g.\ memory or physics). The \textit{Perception Test}~\cite{patraucean2023perception} is the first benchmark to use real world videos and focus on comprehensively diagnosing perception capabilities, like memory, understanding of intuitive physics and geometry, abstractions, or semantics. The benchmark consists of 11.6k videos, with audio, up to 35s long, filmed by diverse crowd-sourced participants following scripts designed to show perceptually-interesting situations. The videos have six types of annotations enabling language and non-language evaluations, across video, audio, and text modalities. The benchmark focuses on probing generalisation and transfer capabilities, so it only provides a relatively small training set to be used for fine-tuning or prompting, and the rest is used for evaluation. The goal of this workshop-challenge was to evaluate SOTA video models using the Perception Test benchmark. More details of the Perception Test and data samples are available on our github repository\footnote{\url{https://github.com/google-deepmind/perception_test}} and workshop website\footnote{\url{https://ptchallenge-workshop.github.io/}}.

\noindent \textbf{Challenge tracks:} Ideally, a single model should be able to perform any evaluation task defined using the annotations provided in the Perception Test. However, such a model does not exist. For this first iteration of the challenge, we defined six individual challenge tracks and evaluate SOTA models whose interface specialises for each annotation type: object tracking, point tracking, temporal action localisation, temporal sound localisation, multiple-choice video QA, grounded video QA.

\noindent \textbf{Challenge setup:} We relied on the open-source eval.ai platform to set up the different challenge tracks. Each track had 2 phases (validation and test), each phase using the corresponding validation and test splits of the Perception Test benchmark. For each submission, the participants had to indicate the evaluation mode (fine-tuning, few-shot, or zero-shot evaluation). In some tracks, the participants had to indicate if the model used the audio modality as well or not (for action and sound localisation, multiple-choice video QA). For test submissions, the participants were required to also upload a short report describing their method (architecture, pre-training datasets and tasks, etc.).
The validation phase served as a sanity check for participants' submission pipelines. The number of submissions for the validation phase was not limited.

The test set was made available 1.5 months before the submission deadline. 
Only the published results on the test set were considered for the competition.  For the test phase, the limit was set to 2 submissions per day, 25 submissions in total. We received 475 submissions from 63 teams across all six tracks in both phases. We awarded 2 prizes per track (best and runner-up) to submissions that obtained the best (and second best) results in the leaderboard. To encourage submissions of novel models, but which are not (yet) competitive with SOTA models, we awarded two \textit{novelty awards} across all tracks; see next section for more details. The reports of the winning submissions are available on the workshop website.

\section{Challenge Tracks, Results, Awards}
In the following we describe each track and the performance achieved in the challenge. For the technical report per team, including winners' affiliations and names, please refer to the workshop's website: \url{https://ptchallenge-workshop.github.io/}.

\subsection{Object tracking}

\noindent \textbf{Task description:} For this task, the model receives a video and a bounding box representing an object, and it is required to track the object throughout the video sequence.

\noindent \textbf{Metric:} The evaluation metric for this task is average Intersection over Union (IoU). It is calculated as the average intersection over union between the predicted bounding boxes and the ground truth bounding boxes for each tracked object.

\noindent \textbf{Dataset:} To make the evaluation task more accessible, we used only a randomly selected subset of 1000 videos from the validation split of the Perception Test for the validation phase, and 1000 videos from the test split of the Perception Test for the test phase; see details about object tracks in Table~\ref{tab:obj_tracks}.

\begin{table}[]
    \centering
    \begin{tabular}{l|r|r}
      \textbf{Split} & \textbf{\# videos} & \textbf{\# object tracks} \\
       \hline
      Train & 2184  & 35373 \\
      Validation  & 1000 & 16501 \\
      Test  & 1000 & 16339 \\
      \hline 
    \end{tabular}
    \caption{Dataset used for the object tracking task.}
    \label{tab:obj_tracks}
\end{table}

\noindent \textbf{Baselines:} We provide two baseline results for this task: a simple dummy baseline, which always assumes that the object is static, i.e. it outputs as predictions the initial bounding box received as input, and a baseline based on SiamFC tracker (UniTrack implementation)~\cite{bertinetto2016fully,wang2021different}.

\begin{figure}
    \centering
    \includegraphics[width=\linewidth]{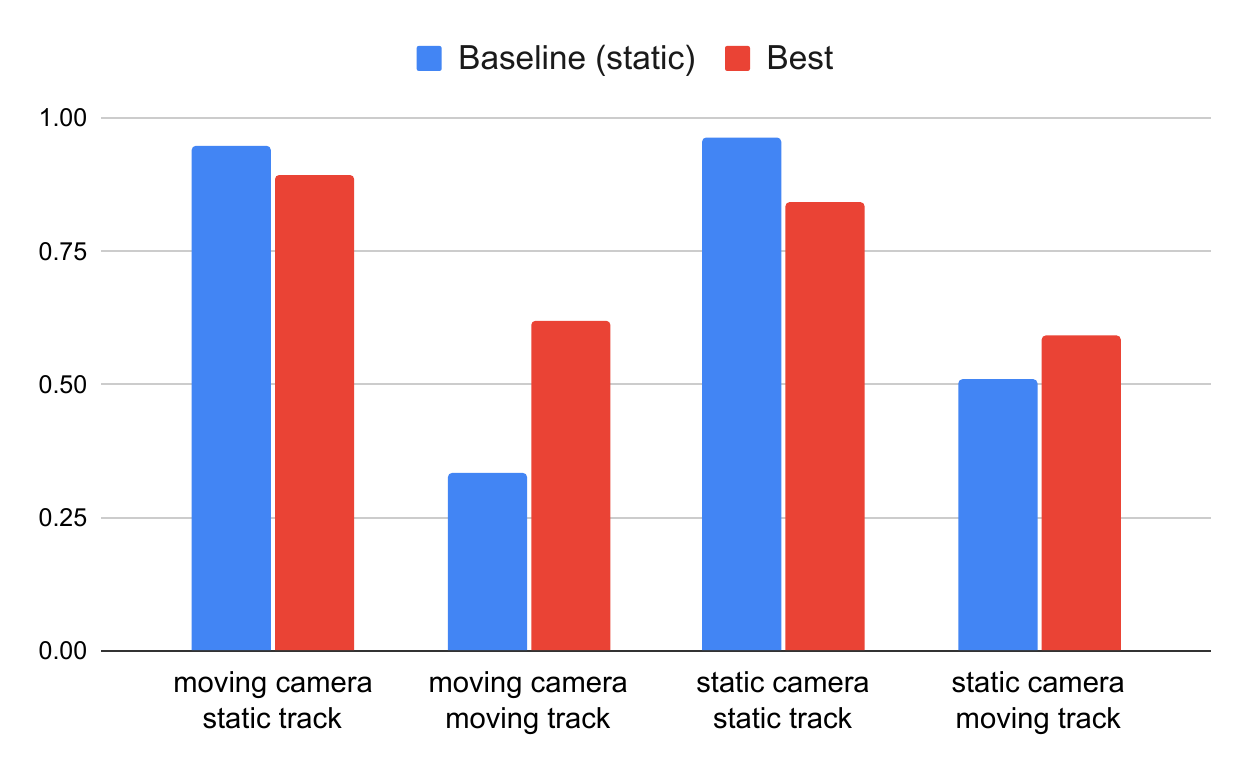}
    \caption{Baseline vs best results split by camera and object motion for the object tracking task.}
    \label{fig:objs}
\end{figure}

\noindent \textbf{Results:}
The results for the top-2 competing models are compared to our baselines in Table~\ref{tab:obj_tracking} and Figure~\ref{fig:objs}. The best performing model, submitted by X-Works Team, relies on the TAT tracker~\cite{he2023target}, incorporating a novel attention module that is able to deal better with long context. Compared to the dummy baseline, this model improves performance significantly on moving objects, when the camera is static or moving.
Check the authors' reports on our workshop page for more details.  

\begin{table}[]
    \centering
    \begin{tabular}{l|l|c}
       \textbf{Rank} & \textbf{Team name} & \textbf{IoU} \\
       \hline
       Baseline 1 & Dummy static & 0.640\\
       Baseline 2 & SiamFC & 0.658 \\
    %   \hline
       Runner-up & sth & 0.732 \\
       Best & X-Works & 0.734 \\
         \hline
    \end{tabular}
    \caption{Object tracking results}
    \label{tab:obj_tracking}
\end{table}

\subsection{Point tracking}

\noindent \textbf{Task description:} In the single point tracking task, the model receives a video and the 2D coordinates of a point, and it is required to track the point throughout the video sequence, also accounting for occlusions.

\noindent \textbf{Metric:} The evaluation metric for this challenge is the average Jaccard, proposed in TAP-Vid~\cite{doersch2022tapvid}. It takes into account 2 scores:

\begin{itemize}
    \item Occlusion Accuracy (OA) -- a simple classification accuracy for the point occlusion prediction on each frame.
    \item Position accuracy -- for frames where the point is visible, it measures the fraction of points that are within a certain threshold of their ground truth. It assumes images are resized to 256x256 pixels. The accuracy is averaged across 5 thresholds: 1, 2, 4, 8, and 16 pixels.
\end{itemize}
The final Jaccard metric calculates the fraction of \textit{true positives}, which are points within the threshold of any visible ground truth points, divided by \textit{true positives} plus \textit{false positives} (points that are predicted as visible but the ground truth is either occluded or farther than the threshold) plus \textit{false negatives} (ground truth visible points that are predicted as occluded or the prediction is farther than the threshold). The overall metric is Jaccard averaged across all thresholds.

\noindent \textbf{Dataset:} We use the subset of videos from the Perception Test that have point tracking annotations; see details in Table~\ref{tab:point_tracks}.
\begin{table}[]
    \centering
    \begin{tabular}{l|r|r}
      \textbf{Split} & \textbf{\# videos} & \textbf{\# point tracks} \\
       \hline
      Train & 28  & 1758 \\
      Validation  & 73 & 4362 \\
      Test  & 44 & 2527 \\
      \hline 
    \end{tabular}
    \caption{Dataset used for the point tracking task.}
    \label{tab:point_tracks}
\end{table}

\noindent \textbf{Baselines:} We provide baseline results for this task using a dummy static baseline, which always assumes that the point is static.

% , and TAP-Net~\cite{doersch2022tapvid}.

\begin{figure}
    \centering
    \includegraphics[width=\linewidth]{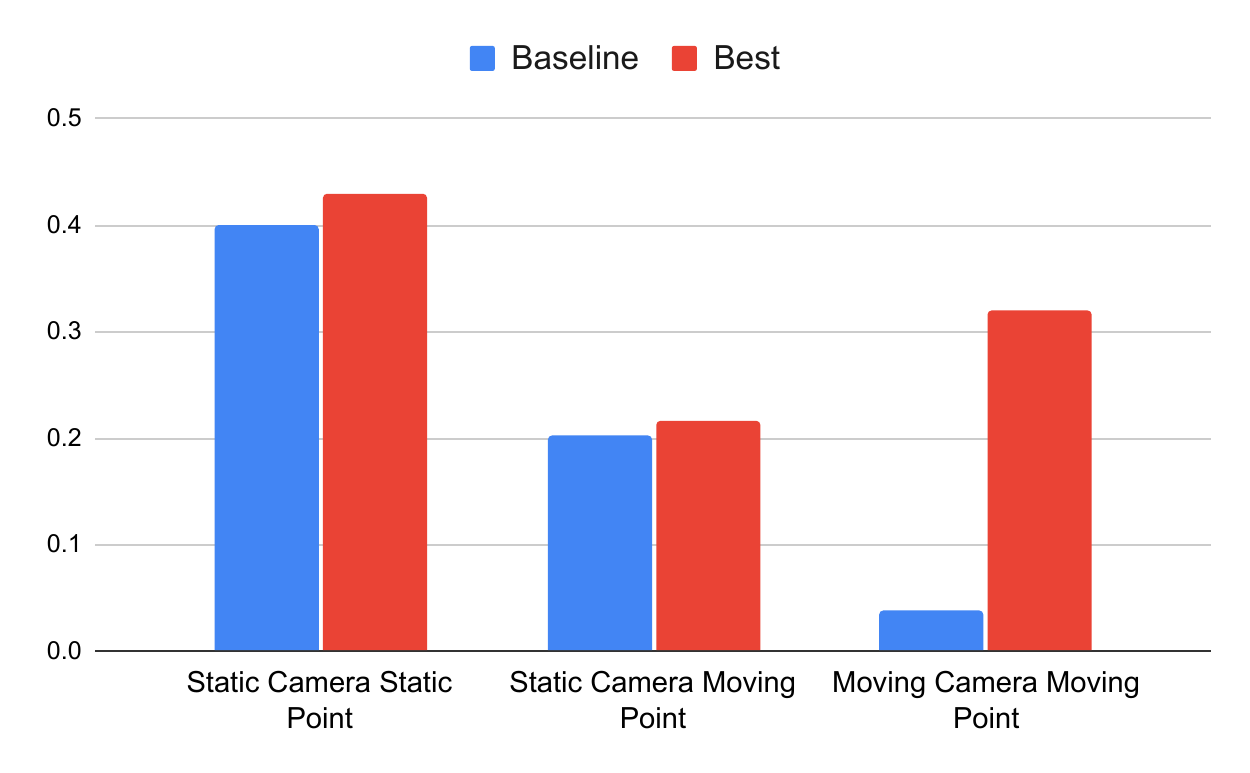}
    \caption{Baseline vs best results split by camera and point motion for the point tracking task.}
    \label{fig:pts}
\end{figure}

\noindent \textbf{Results:}
Table~\ref{tab:point_tracking} and Figure~\ref{fig:pts} show the results of the top-2 competing models compared to our static dummy baseline. The best results were obtained by NJUST\_KMG\_Point Team, using a TAPIR-based model~\cite{doersch2023tapir}, and show a significant improvement over the baseline especially for moving points. This is due to explicitly modelling the camera motion and segmenting the moving regions in the video; please check the workshop website for more details on the method included in the submission report.

\begin{table}[]
    \centering
    \begin{tabular}{l|l|c}
       \textbf{Rank} & \textbf{Team name} & \textbf{Jaccard} \\
       \hline
       Baseline & Dummy static & 0.418 \\
    %   \hline
       Runner-up & THETEAM & 0.451 \\
       Best & NJUST\_KMG\_Point & 0.458 \\
         \hline
    \end{tabular}
    \caption{Point tracking results}
    \label{tab:point_tracking}
\end{table}

\subsection{Temporal action localisation}

\noindent \textbf{Task description:} In the temporal action localisation task, the model receives a video and is required to localise and classify the actions occurring in the video according to a predefined set of classes; there are 63 action classes in total.

\noindent \textbf{Metric:} The evaluation metric for this challenge is mean average precision (mAP). It is calculated as the average precision over different action classes and IoU thresholds. For the IoU thresholds in evaluation we use [0.1 $\rightarrow$ 0.5] with 0.1 increments, similar to~\cite{Damen2021TheED}.

\noindent \textbf{Dataset:} We use the videos from the Perception Test for this challenge. To facilitate experimentation, we also provide features for the video / audio modalities that participants could optionally use for their submissions: video features extracted using TSP~\cite{alwassel2021tsp} and audio features extracted using MMV~\cite{alayrac2020self}. Each video has multiple action segment annotations; see details in Table~\ref{tab:tal_dataset}.

\begin{table}[]
    \centering
    \begin{tabular}{l|r|r}
      \textbf{Split} & \textbf{\# videos} & \textbf{\# action segments} \\
       \hline
      Train & 2009  & 13097 \\
      Validation  & 5359 & 35440 \\
      Test  & 3233 & 20741 \\
      \hline 
    \end{tabular}
    \caption{Dataset used for the temporal action localisation task.}
    \label{tab:tal_dataset}
\end{table}

\noindent \textbf{Baselines:} The baseline for this task is ActionFormer~\cite{zhang2022actionformer}, a state-of-the-art model for temporal action localisation, that we fine-tuned for the set of classes present in our benchmark.

\begin{figure*}
    \centering
    \includegraphics[width=.48\linewidth]{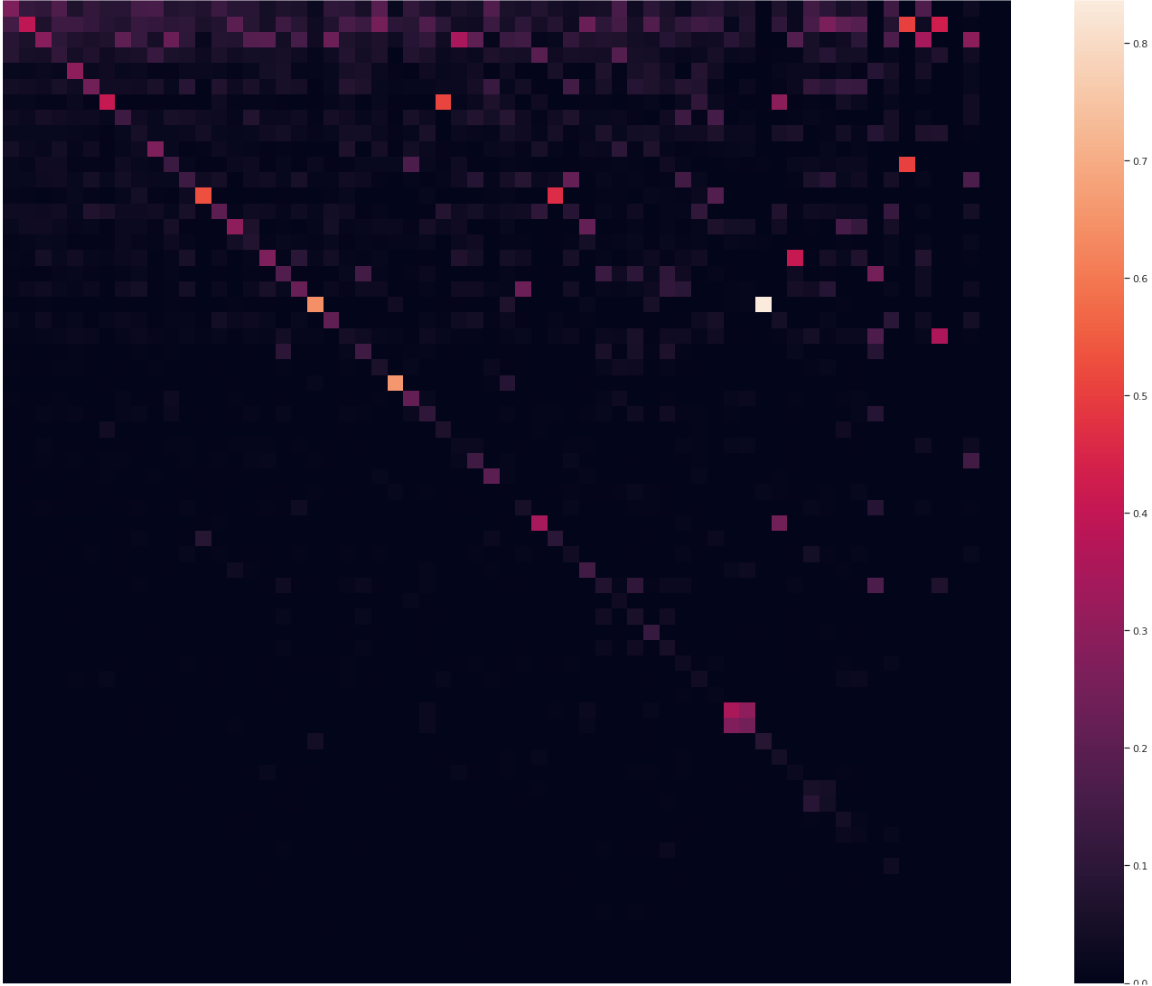}
    \includegraphics[width=.48\linewidth]{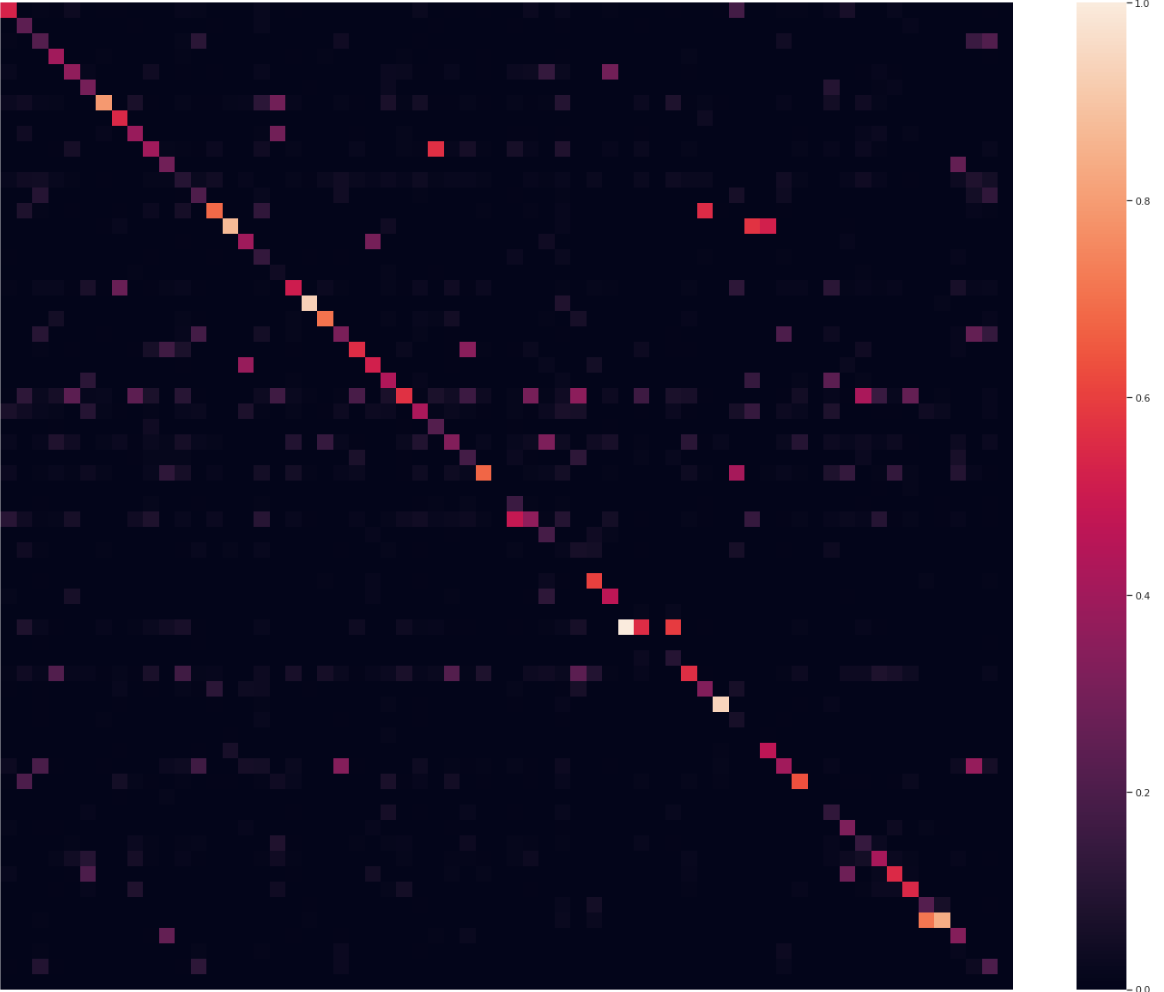}
    \caption{Baseline (left) vs best results (right) for the temporal action localisation task.}
    \label{fig:tal}
\end{figure*}

\noindent \textbf{Results:} 
The results of the top-2 competing methods are included in Table~\ref{tab:tal} and Figure~\ref{fig:tal}. It can be observed that they improve performance significantly over the ActionFormer baseline, across all classes. The top entry, submitted by CTCV Team, relied on the same model as the baseline (ActionFormer), but used features extracted using pre-trained ViT models~\cite{dosovitskiy2020vit}, plus a Weighted Boxes Fusion operation~\cite{solovyev2021weighted} instead of the standard non-maxima suppression. In addition, the authors reported filming additional videos for 3 classes that were poorly represented in the training set. Please check the authors' report on our workshop page for more details.   

\begin{table}[]
    \centering
    \begin{tabular}{l|l|c}
       \textbf{Rank} & \textbf{Team name} & \textbf{mAP} \\
       \hline
       Baseline & ActionFormer & 0.156\\
    %   \hline
       Runner-up & OpenGVLab & 0.465\\
       Best & CTCV & 0.504 \\
         \hline
    \end{tabular}
    \caption{Temporal action localisation results}
    \label{tab:tal}
\end{table}

\subsection{Temporal sound localisation}

\noindent \textbf{Task description:} In the temporal sound localisation task, the model receives a video and is required to localise and classify the sound events occurring in the video according to a predefined set of sound classes; there are 16 sound classes in our dataset. For the challenge, we consider only 12 classes, excluding classes like \textit{Background}, \textit{Background-Other}, \textit{Human-Other}, \textit{Animal-Other} due to their ambiguity. 

\noindent \textbf{Metric:} Similar to the action localisation task above, the metric for this challenge is mean average precision (mAP). It is calculated as the average precision over different sound classes and IoU thresholds. For the IoU thresholds in evaluation we use [0.1 $\rightarrow$ 0.5] with 0.1 increments.

\noindent \textbf{Dataset:} As for the temporal action localisation task above, we provide the same features for all the videos in the Perception Test; see Table~\ref{tab:sounds} for details.

\begin{table}[]
    \centering
    \begin{tabular}{l|r|r}
      \textbf{Split} & \textbf{\# videos} & \textbf{\# sound segments} \\
       \hline
      Train & 2009  & 13289 \\
      Validation  & 5359 & 35625 \\
      Test  & 3238 & 21636 \\
      \hline 
    \end{tabular}
    \caption{Dataset used for the temporal action localisation task.}
    \label{tab:sounds}
\end{table}

\noindent \textbf{Baselines:} We provide baseline results for this task using the same model as in the action localisation task ActionFormer~\cite{zhang2022actionformer}, adapted to the sound localisation task by fine-tuning on our sound annotations belonging to the train split.

\begin{figure}
    \centering
    \includegraphics[width=.48\linewidth]{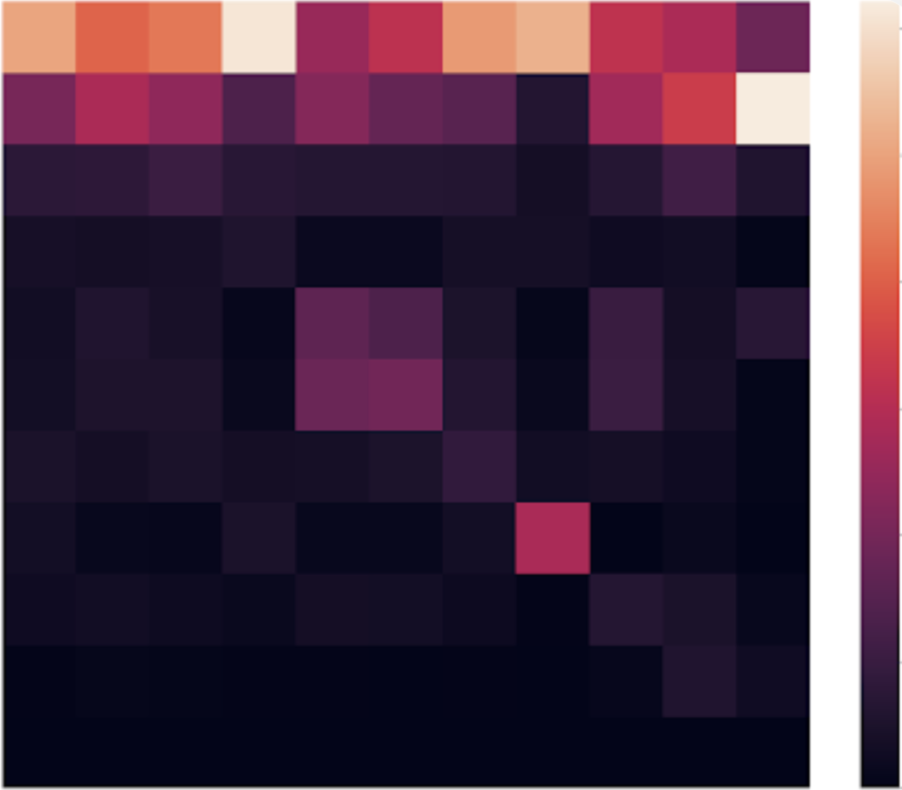}
    \includegraphics[width=.48\linewidth]{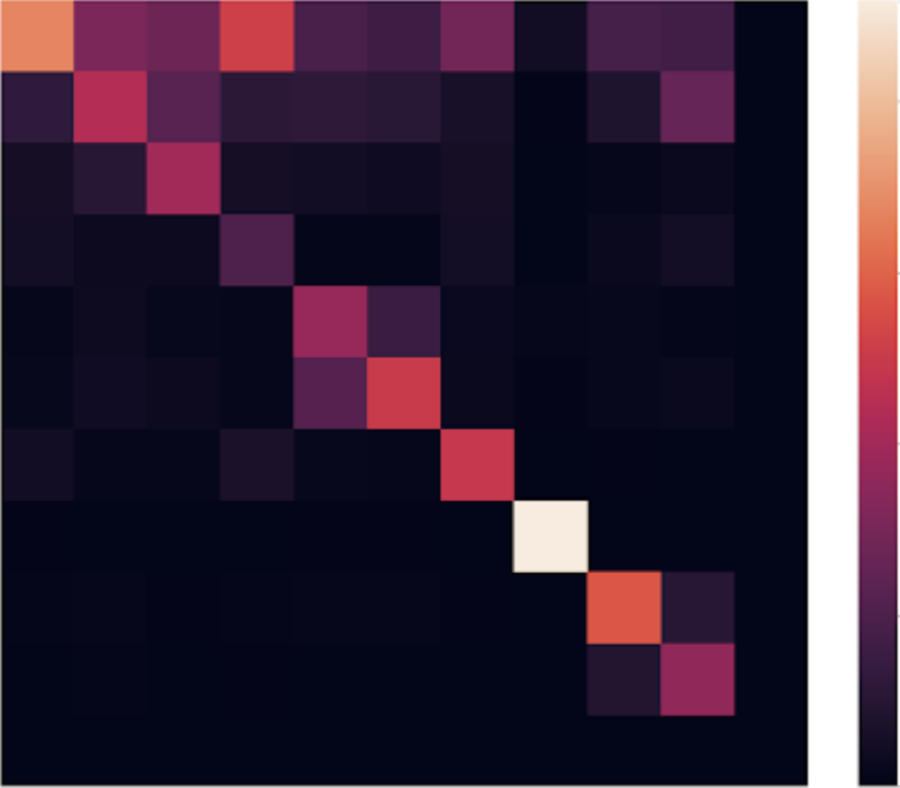}
    \caption{Baseline (left) vs best results (right) for the temporal sound localisation task.}
    \label{fig:tsl}
\end{figure}

\noindent \textbf{Results:}
Table~\ref{tab:tsl} and Figure~\ref{fig:tsl} show the performance of the top-2 competing methods in this track, compared to our baseline (ActionFormer). Both methods use multimodal inputs and yield much better performance compared to our baseline across all sound classes. The top entry, submitted by OpenGVLab Team, relied on ActionFormer with video features extracted using UMT-Large~\cite{li2023unmasked} and audio features extracted using BEAT~\cite{pmlr-v202-chen23ag}; see authors' report included on our workshop page for more details.     

\begin{table}[]
    \centering
    \begin{tabular}{l|l|c}
       \textbf{Rank} & \textbf{Team name} & \textbf{mAP} \\
       \hline
       Baseline & ActionFormer & 0.102 \\
    %   \hline
       Runner-up & NJUST\_KMG & 0.331 \\
       Best & OpenGVLab & 0.400 \\
         \hline
    \end{tabular}
    \caption{Temporal sound localisation results.}
    \label{tab:tsl}
\end{table}

\subsection{Multiple-choice video QA}

\noindent \textbf{Task description:} In the multiple-choice video question-answering (mc-vQA) task, the model receives, in parallel with the video, a question and three possible answers, out of which only one is correct, and the model has to pick one answer. The questions cover four skill areas (Memory, Abstraction, Physics, Semantics) and require different types of reasoning (Descriptive, Explanatory, Predictive, Counterfactual), across video, audio, and text modalities. The questions are also tagged with skills in each area such as: event recall (Memory), object counting (Abstraction), collision (Physics), action recognition (Semantics) and more.

\noindent \textbf{Metric:} The evaluation metric for this challenge is top-1 accuracy. It is calculated as the percentage of questions where the model's predicted option id (1 out of 3) matches the ground truth option id.

\noindent \textbf{Dataset:} Each video in the dataset has a number of multiple-choice video QA tasks associated, each question having 3 options, out of which only one is correct; see Table~\ref{tab:mcqa-dataset} for details.
\begin{table}[]
    \centering
    \begin{tabular}{l|r|r}
      \textbf{Split} & \textbf{\# videos} & \textbf{\# questions} \\
       \hline
      Train & 2184  & 7392 \\
      Validation  & 5900 & 19140 \\
      Test  & 3525 & 11528 \\
      \hline 
    \end{tabular}
    \caption{Dataset used for the multiple-choice video QA task.}
    \label{tab:mcqa-dataset}
\end{table}

\noindent \textbf{Baselines:} We provide baseline results for this task using a dummy frequency-based baseline, with multiple setups: 0-shot, few-shot, all-shot.

\begin{figure*}[t]
    \centering
    \includegraphics[width=.48\linewidth]{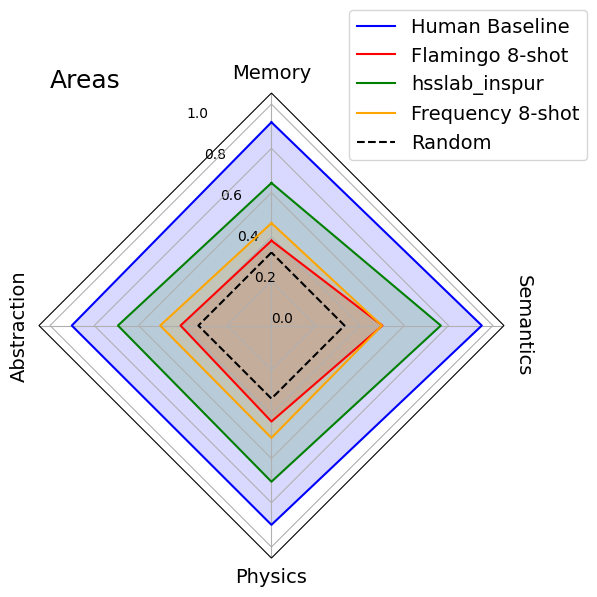}
    \includegraphics[width=.43\linewidth]{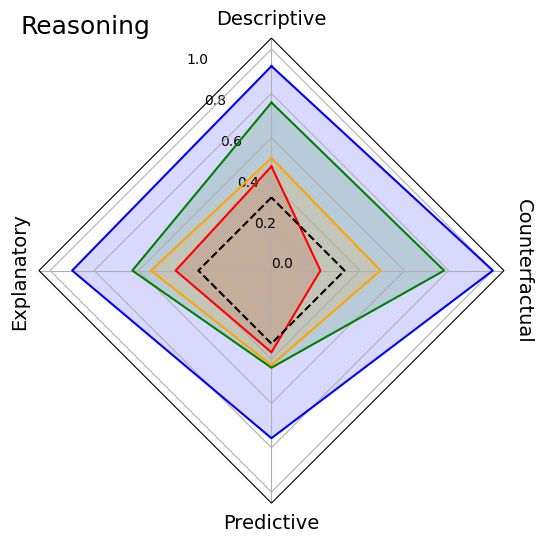}
    \caption{Baselines vs best and runner up models detailed by areas and types of reasoning for the multiple-choice video QA task.}
    \label{fig:mcqa}
\end{figure*}

\begin{figure*}[t]
    \centering
    \includegraphics[width=\linewidth]{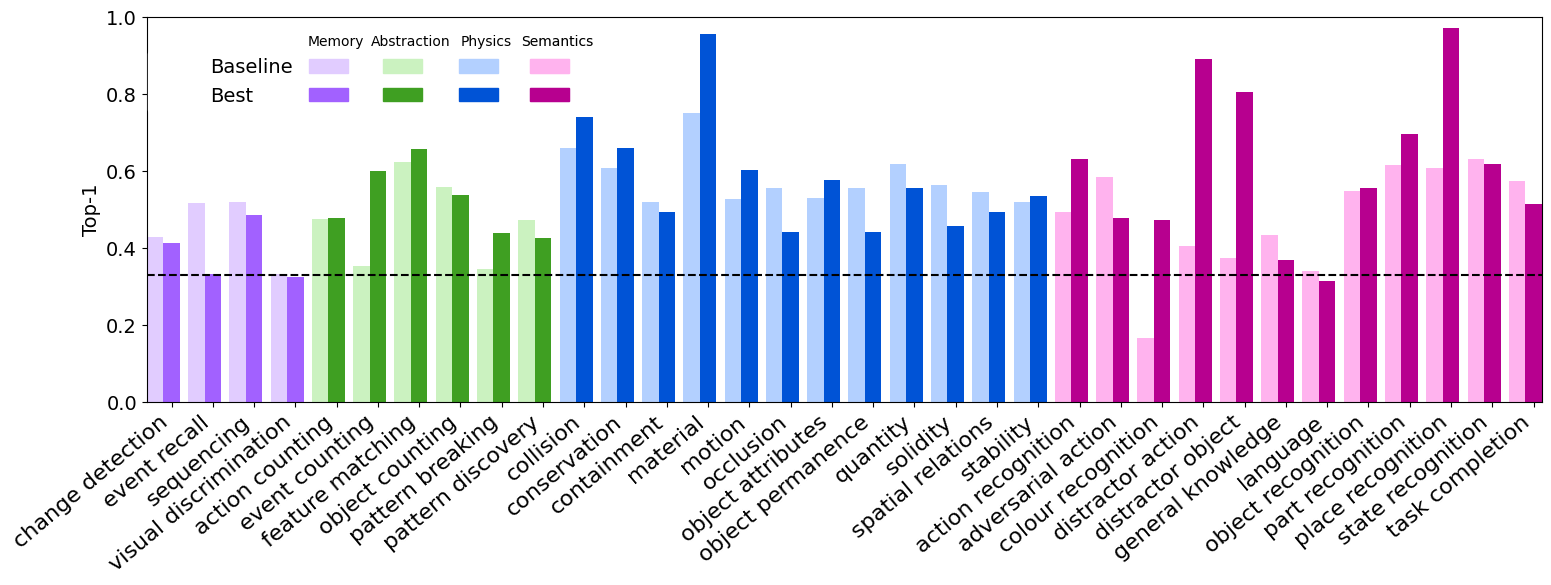}
    \caption{Baseline vs best model detailed by skill, and grouped by area, for the multiple-choice video QA task.}
    \label{fig:mcqa2}
\end{figure*}

\noindent \textbf{Results:}
Table~\ref{tab:mcqa} and Figure~\ref{fig:mcqa} show the performance of the top-2 competing models compared to our baselines (including a human one). Please check our paper for details about the frequency and the human baselines included in our benchmark.
Both top-2 competing models relied on the same model as our SeViLA ~\cite{yu2023self} baseline. The top entry, submitted by hsslab\_inspur Team, replaced the video and text components in SeViLA for larger models (Video-LLAMA~\cite{damonlpsg2023videollama} and Vicuna-13B~\cite{vicuna2023} respectively), pre-trained on multiple external datasets and fine-tuned on the training split of our benchmark; see the authors' report on our workshop webpage for more details. 

We provide a detailed comparison by individual skills in Fig~\ref{fig:mcqa2}.
Through fine-tuning, the performance improved mainly in the semantics area (for skills like recognition of distractor actions and objects), but in Memory or Physics area, the performance does not exceed the dummy frequency baseline significantly. Overall, the performance is still far from the human baseline, which is obtained zero-shot.          

\begin{table}[]
    \centering
    \begin{tabular}{l|l|c}
       \textbf{Rank} & \textbf{Team name} & \textbf{top-1} \\
       \hline
       Baseline 1 & Frequency (0-shot) & 0.335 \\
       Baseline 2 & Frequency (8-shot) & 0.510 \\
       Baseline 3 & Frequency (all-shot) & 0.552 \\
    %   \hline
       Runner-up & TTgogogo (fine-tuned) & 0.683 \\
       Best & hsslab\_inspur (fine-tuned) & 0.715 \\
         \hline
    \end{tabular}
    \caption{Multiple-choice video QA results.}
    \label{tab:mcqa}
\end{table}

\subsection{Grounded video QA}

\noindent \textbf{Task description:} In the grounded video QA task, the model receives a video and a question/query as input, and it is required to track throughout the video the object(s) that represent the answer to the question. This is a novel type of grounded video QA task.

\noindent \textbf{Metric:} The evaluation metric for this challenge is HOTA (Higher Order Tracking Accuracy)~\cite{luiten2020IJCV}. It unifies the detection, association, and localization accuracy into a single metric.

\noindent \textbf{Dataset:} We use the videos from the Perception Test that have annotations for this task; see Table~\ref{tab:gqa-dataset} for details.
\begin{table}[]
    \centering
    \begin{tabular}{l|r|r}
      \textbf{Split} & \textbf{\# videos} & \textbf{\# questions} \\
       \hline
      Train & 586  & 1859 \\
      Validation  & 1545 & 3051 \\
      Test  & 932 & 1859 \\
      \hline 
    \end{tabular}
    \caption{Dataset used for the grounded video QA task.}
    \label{tab:gqa-dataset}
\end{table}

\noindent \textbf{Baselines:} Since this is a novel task, no baseline exists in the literature. We provide a simple baseline, which runs MDETR detector~\cite{kamath2021mdetr} on the middle frame of the video using the given question as query, then it keeps the detections static throughout the video.

\begin{figure}
    \centering
    \includegraphics[width=\linewidth]{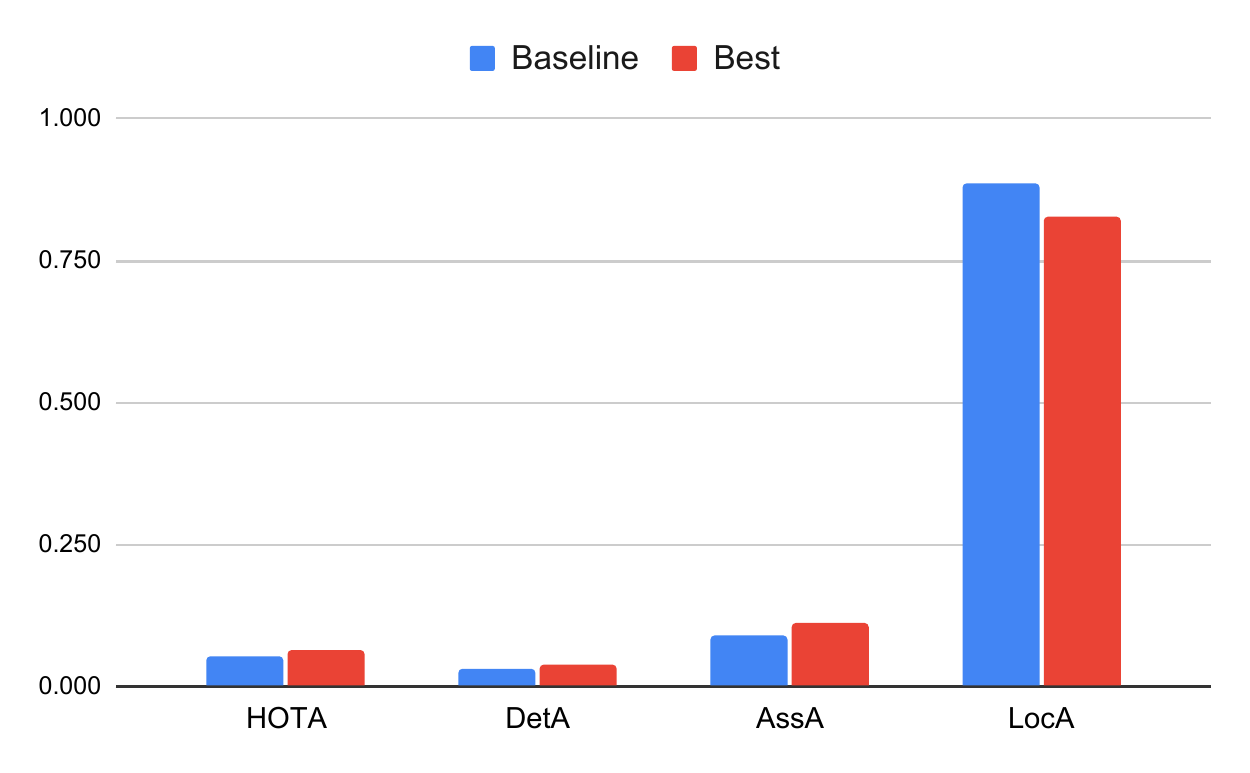}
    \caption{Baseline vs best results in terms of overall HOTA, detection, assignment, and localisation accuracy for the grounded video QA task.}
    \label{fig:hota}
\end{figure}

\noindent \textbf{Results:}
In this track, we received submissions that exceeded the performance of the baseline from a single team (NJUST--KMG), hence we did not award a runner-up prize. The results are included in Table~\ref{tab:gqa} and Figure~\ref{fig:hota}, compared to our baseline. The method relies on VALOR~\cite{chen2023valor} to get text answers for the given video-question pairs and then uses TubeDETR~\cite{yang2022tubedetr} to ground the answers. The results are still fairly poor, showing that this is a very challenging task for existing models.

\begin{table}[]
    \centering
    \begin{tabular}{l|l|c}
       \textbf{Rank} & \textbf{Team name} & \textbf{HOTA} \\
       \hline
       Baseline & MDETR+static & 0.057 \\
    %   \hline
              Runner-up & Not awarded & - \\
       Best & NJUST--KMG & 0.063 \\
         \hline
    \end{tabular}
    \caption{Grounded video question-answering results.}
    \label{tab:gqa}
\end{table}

\subsection{Novelty awards}
The organising team selected 2 most novel entries across all tracks; see Table~\ref{tab:novelty}. The point tracking entry relies on dense feature matching across frames using LoFTR~\cite{sun2021loftr}, obtaining competitive results (runner-up in point tracking). Inspired by FocalNet~\cite{yang2022focal}, the sound localisation entry  proposed a Focal Audio-Visual Network for both action and sound localisation, leading to competitive results in the audio track (3rd place). Please check the corresponding submission reports on our workshop webpage for more details.   

\begin{table}[]
    \centering
    \begin{tabular}{l|l}
       \textbf{Team name} & \textbf{Award} \\
       \hline
       THETEAM & Novel point tracking \\
       JNU\_boat & Novel sound localisation \\
         \hline
    \end{tabular}
    \caption{Grounded video question-answering results.}
    \label{tab:novelty}
\end{table}

\section{Discussion}

The first Perception Test challenge attracted a lot of interest, reflected in the large number of submissions and participating teams. The submitted models improved the results over the provided baselines in each track; please check the corresponding evaluation webpages for comparisons with submitted models. However, there is still significant room for improvement in each track -- the benchmark is nowhere near saturation. In addition, none of the submitted models was evaluated on multiple tracks. In our next iteration, we plan to explicitly incentivise participants to use a single model for multiple tasks (through awards or imposed), and restrict the evaluation modes to zero-shot/few-shot (no fine-tuning) to robustly test for generalisation capabilities. 

\subsection*{Acknowledgements} We would like to thank Relja Arandjelovic and Volodymyr Mnih for reviewing this report. We are grateful to Google DeepMind for providing the funding for the awards, and to Hayley Delew and James Robson from Google DeepMind and Ludivine Fluneau and Francois Tapissier from Dakini for ensuring a smooth handling of the awards. Special thanks to Gunjan Chablani from Eval AI for support and additional resources while running the challenges.

\bibliography{main}

\end{document}